\documentclass{article} %
\usepackage{colm2024_conference}

\usepackage{booktabs}
\usepackage{graphicx}
\usepackage{enumitem}
\usepackage{wrapfig}
\usepackage{algorithm}
\usepackage{algpseudocode}

\usepackage{microtype}
\usepackage{amsmath}
\usepackage{amsfonts}
\usepackage{colortbl}
\usepackage[utf8]{inputenc}
\usepackage[T1]{fontenc}
\definecolor{lightgray}{rgb}{0.9,0.9,0.9}
\usepackage{caption}
\usepackage{subcaption}
\usepackage{xcolor}
\usepackage{setspace}
\usepackage{url}
\usepackage{multirow}
\usepackage{colortbl}
\usepackage{tabularx}
\usepackage{blindtext}
\usepackage{pgfplots}
\pgfplotsset{compat=1.18} 
\usepackage{tikz}
\usetikzlibrary{er,positioning,bayesnet}
\usepackage{makecell}
\usepackage{tipa}
\usepackage{siunitx}
\usepackage{nicefrac}
\usepackage{tocloft}
\usepackage{listings}
\usepackage[raster,skins]{tcolorbox} %
\usepackage{xltabular}
\usepackage{adjustbox}
\usepackage{xurl}
\usepackage{xcolor}

\usepackage{amsmath,amsfonts,bm}

\def\eqref#1{equation~\ref{#1}}

\def\1{\bm{1}}

\DeclareMathAlphabet{\mathsfit}{\encodingdefault}{\sfdefault}{m}{sl}
\SetMathAlphabet{\mathsfit}{bold}{\encodingdefault}{\sfdefault}{bx}{n}

\title{Qwen2.5-Omni Technical Report}

\author{
\bf Qwen Team
}

\newcommand{\method}{Qwen2.5-Omni\xspace}

\begin{document}

\maketitle

\begin{abstract}
In this report, we present \method, an end-to-end multimodal model designed to perceive diverse modalities, including text, images, audio, and video, while simultaneously generating text and natural speech responses in a streaming manner. 
To enable the streaming of multimodal information inputs,  both audio and visual encoders utilize a block-wise processing approach. This strategy effectively decouples the handling of long sequences of multimodal data, assigning the perceptual responsibilities to the multimodal encoder and entrusting the modeling of extended sequences to a large language model. Such a division of labor enhances the fusion of different modalities via the shared attention mechanism. 
To synchronize the timestamps of video inputs with audio, we organize the audio and video sequentially in an interleaved manner and propose a novel position embedding approach, named \textbf{TMRoPE}~(\textbf{T}ime-aligned \textbf{M}ultimodal \textbf{RoPE}).
To concurrently generate text and speech while avoiding interference between the two modalities, we propose \textbf{Thinker-Talker} architecture. In this framework, Thinker functions as a large language model tasked with text generation, while Talker is a dual-track autoregressive model that directly utilizes the hidden representations from the Thinker to produce audio tokens as output. Both the Thinker and Talker models are designed to be trained and inferred in an end-to-end manner. 
For decoding audio tokens in a streaming manner, we introduce a sliding-window DiT that restricts the receptive field, aiming to reduce the initial package delay.
\method~is comparable with similarly sized Qwen2.5-VL and outperforms Qwen2-Audio. Furthermore, \method~achieves state-of-the-art performance on multimodal benchmarks like Omni-Bench. Notably, \method~'s performance in end-to-end speech instruction following is comparable to its capabilities with text inputs, as evidenced by benchmarks such as MMLU and GSM8K. As for speech generation, \method's streaming Talker outperforms most existing streaming and non-streaming alternatives in robustness and naturalness.


\end{abstract}





\begin{figure}[tbh]
    \centering
    \includegraphics[width=0.8\textwidth]{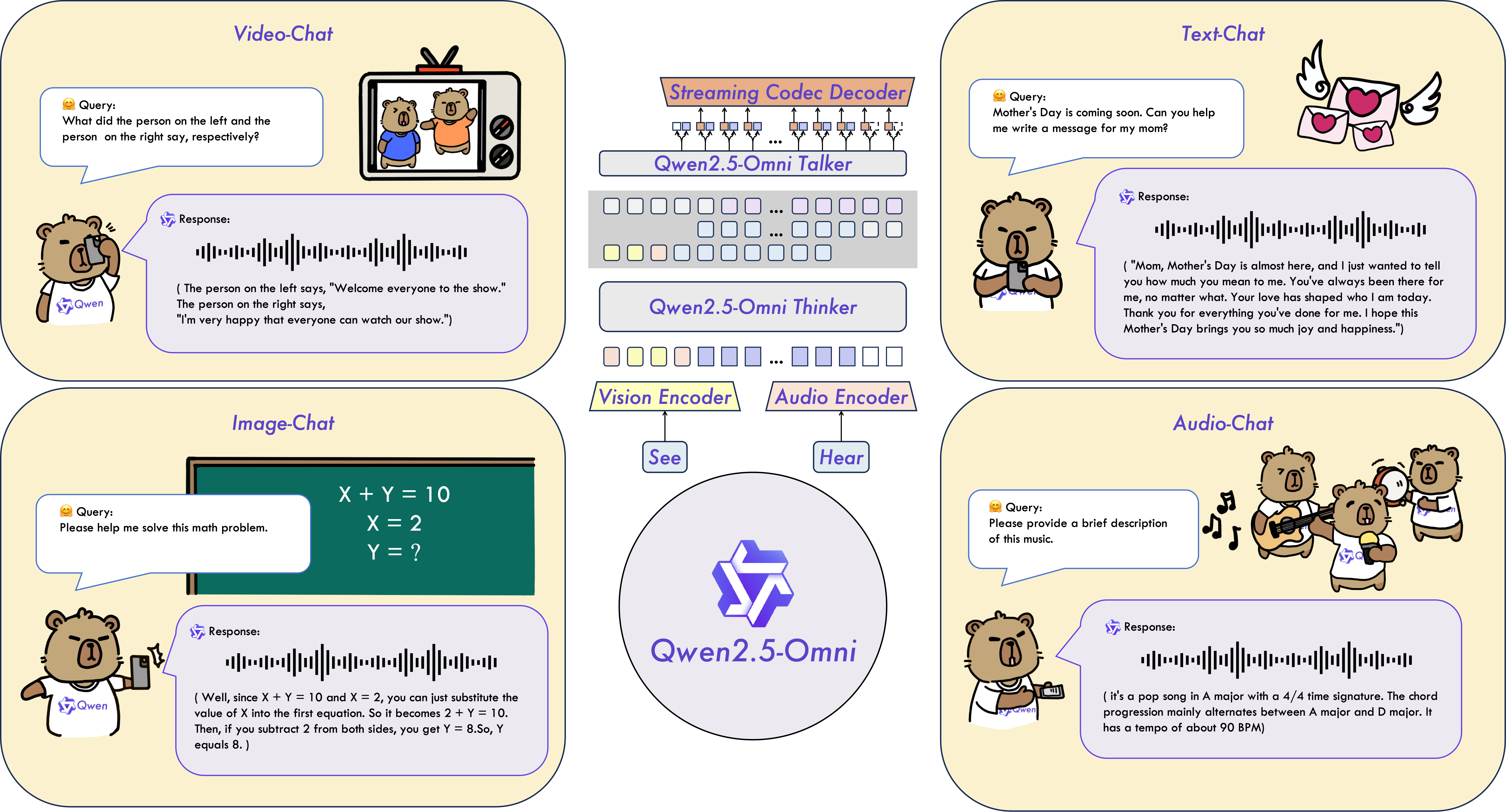}
    \caption{\method~is a unified end-to-end model capable of processing multiple modalities, such as text, audio, image and video, and generating real-time text or speech response. Based on these features, \method~supports a wide range of tasks, including but not limited to voice dialogue, video dialogue, and video reasoning.}
    \label{fig:ovewview_arch}
\end{figure}

\section{Introduction}
\label{sec:intro}

In daily life, humans are capable of simultaneously perceiving the visual and auditory information around them. After processing this information through the brain, they express feedback through writing, vocalization, or using tools (and physical actions), thereby engaging in information exchange with various organisms in the world and exhibiting intelligence. In recent years, general artificial intelligence has become increasingly visible, largely due to advancements in Large Language Models (LLMs)~\citep{gpt3,gpt4,gpt4o,gemini,claude,claude2,claude3,qwen,qwen2,llama,llama2,llama3}. These models, trained on vast amounts of textual data, represent high-level discrete representation created by humans, showcasing the ability to solve complex problems and learn rapidly. Furthermore, in the realm of understanding, Language-Audio-Language Models (LALMs)~\citep{gpt4o,anonymous2023salmonn,qwenaudio, qwen2-audio} and Language-Visual-Language Models (LVLMs)~\citep{blip2,llava,instructblip,minigpt-4,kosmos,qwenvl,llava1.5,cogvlm,gpt4v,gemini} have helped LLMs to further extend auditory and visual capabilities in an end-to-end manner. 
However, efficiently unifying all these different understanding modalities in an end-to-end fashion, utilizing as much data as possible, and providing responses in both text and speech streams akin to human communication still presents a significant challenge.

The development of a unified and intelligent omni-model requires careful consideration of several key factors. First, it is crucial to implement a systematic method for the joint training of various modalities, including text, images, videos, and audio, to foster mutual enhancement among them. This alignment is particularly important for video content, where synchronization of the temporal aspects of audio and visual signals is necessary. Second, it is essential to manage potential interference among outputs from different modalities, ensuring that the training processes for outputs such as text and voice tokens do not disrupt each other. Finally, there is a need to explore architectural designs that enable real-time understanding of multimodal information and allow for efficient audio output streaming, thereby reducing initial latency.

In this report, we introduce \method, a unified single model capable of processing multiple modalities and generating text and natural speech responses simultaneously in a streaming format. To tackle the first challenge, we propose a novel position embedding approach, named \textbf{TMRoPE}~(\textbf{T}ime-aligned \textbf{M}ultimodal \textbf{RoPE}). We organize these audio and video frames in an interleaved structure to represent video sequences in time order. 
For the second challenge, we present Thinker-Talker architecture, wherein Thinker is tasked with text generation while the Talker focuses on generating streaming speech tokens. Talker receives high-level representations directly from Thinker. This design is inspired by the way humans utilize different organs to produce various signals, which are simultaneously coordinated through the same neural networks. As a result, Thinker-Talker architecture is end-to-end jointly trained, with each component dedicated to generating distinct signals.
To address the challenges associated with streaming and to facilitate the pre-filling necessary for real-time comprehension of multimodal signals, we propose modifications to all multimodal encoders by adopting a block-wise streaming processing approach. In order to support streaming speech generation, we implement a dual-track autoregressive model that generates speech tokens, alongside a DiT model which converts these tokens into waveforms, thereby enabling streaming audio generation and minimizing initial latency. This design aims to enable the model to process multimodal information in real-time and effectively perform pre-filling, thereby enabling the concurrent generation of text and speech signals. 

\method is comparable with the similarly sized Qwen2.5-VL~\citep{qwen2vl} and outperforms Qwen2-Audio~\citep{qwen2-audio} in image and audio capabilities respectively. Furthermore, \method achieves state-of-the-art performance on multimodal benchmarks such as OmniBench~\citep{li2024omnibench} and AV-Odyssey Bench~\citep{gong2024av}. Notably, \method's performance in end-to-end speech instruction following is comparable to its capabilities with text inputs, as evidenced by benchmarks such as MMLU~\citep{mmlu} and GSM8K~\citep{gsm8k}. As for speech generation, \method achieves 1.42\%, 2.33\% and 6.54\% WER on seed-tts-eval~\citep{seedtts} test-zh, test-en and test-hard set respectively, outperforming MaskGCT~\citep{maskgct} and CosyVoice~2~\citep{cosyvoice2}.

The key features of \method can be summarized as:

\begin{itemize}
    \item  We introduce \method, a unified model that can perceive all modalities and simultaneously generate text and natural speech responses in a streaming fashion.
    \item We present a novel positional embedding algorithm, termed TMRoPE, which explicitly incorporates temporal information for synchronizing audio and video.
    \item We propose the Thinker-Talker Architecture to facilitate real-time comprehension and speech generation.
    \item \method demonstrates strong performance across all modalities when benchmarked against similarly sized single-modality models. It significantly enhances the capability of following voice commands, achieving performance levels comparable to pure text input. For tasks that involve integrating multiple modalities, such as those evaluated in OmniBench~\citep{li2024omnibench}, \method achieves state-of-the-art performance. Notably, \method achieves strong performance on seed-tts-eval~\citep{seedtts}, demonstrating robust speech generation abilities.
\end{itemize}

\section{Architecture}
\begin{figure}[tbh]
    \centering
    \includegraphics[width=0.8\textwidth]{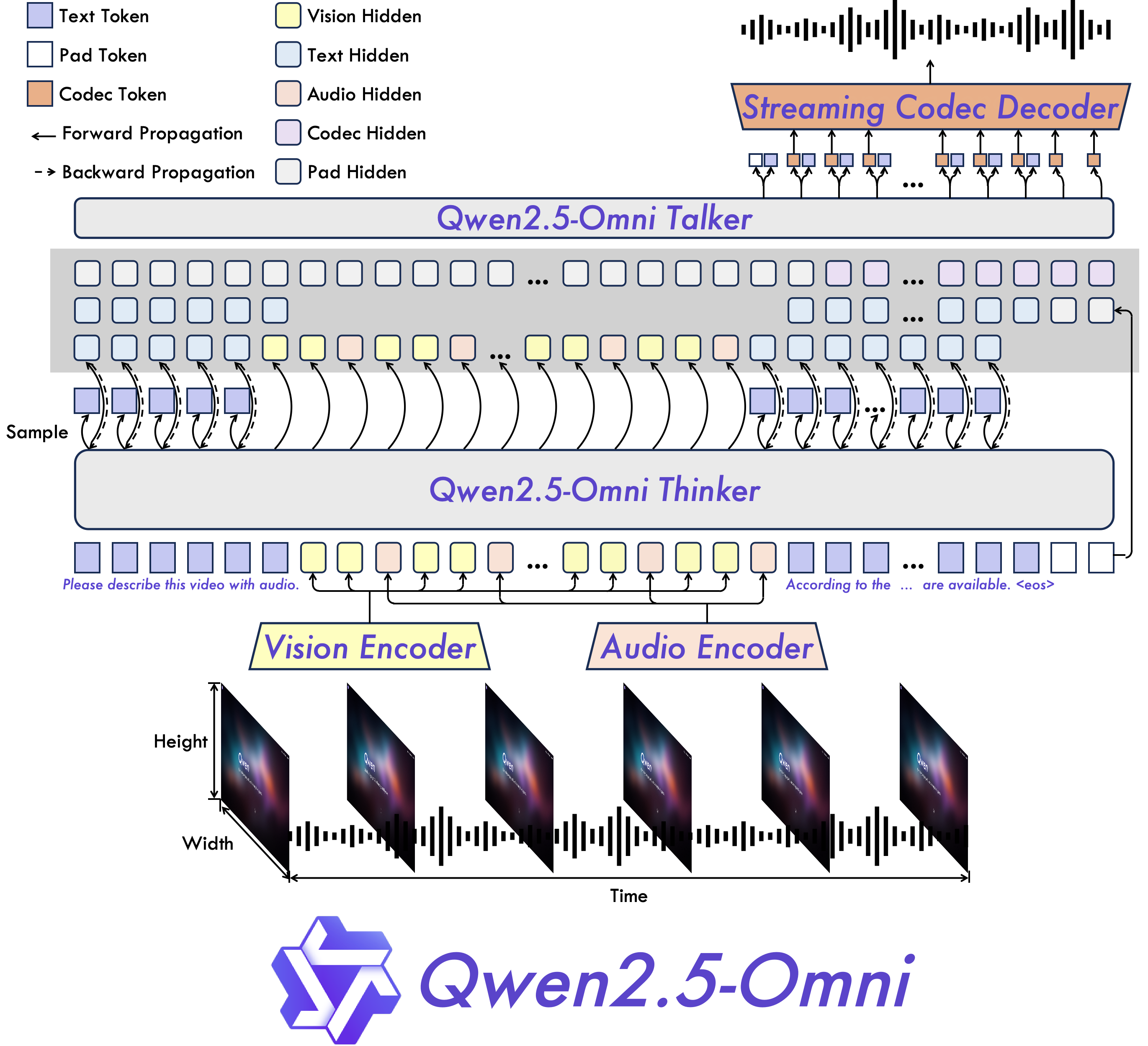}
    \caption{The overview of \method. \method adpots the Thinker-Talker architecture. Thinker is tasked with text generation while Talker focuses on generating streaming speech tokens by receives high-level representations directly from Thinker.}
    \label{fig:ovewview_arch}
\end{figure}

\subsection{Overview}
As shown in Figure~\ref{fig:ovewview_arch}, \method employs Thinker-Talker architecture. Thinker functions like a brain, responsible for processing and understanding inputs from text, audio and video modalities, generating high-level representations and corresponding text. Talker operates like a human mouth, taking in the high-level representations and text produced by the Thinker in a streaming manner, and outputting discrete tokens of speech fluidly. Thinker is a Transformer decoder, accompanied by encoders for audio and image that facilitate information extraction. In contrast, Talker is designed as a dual-track autoregressive Transformer Decoder architecture, motivated by Mini-Omni~\citep{xie2024mini}. During both training and inference, Talker directly receives high-dimensional representations from Thinker and shares all of Thinker's historical context information. Consequently, the entire architecture operates as a cohesive single model, enabling end-to-end training and inference.

In the following sections, we first introduce how \method perceives various input signals and present our proposed novel positional encoding algorithm, TMRoPE. Subsequently, the details of text and speech generation are presented. Finally, we highlight the improvements made in the understanding and generation modules to facilitate efficient streaming inference.

\subsection{Perceivation}
\begin{figure}[tbh]
    \centering
    \includegraphics[width=1.0\textwidth]{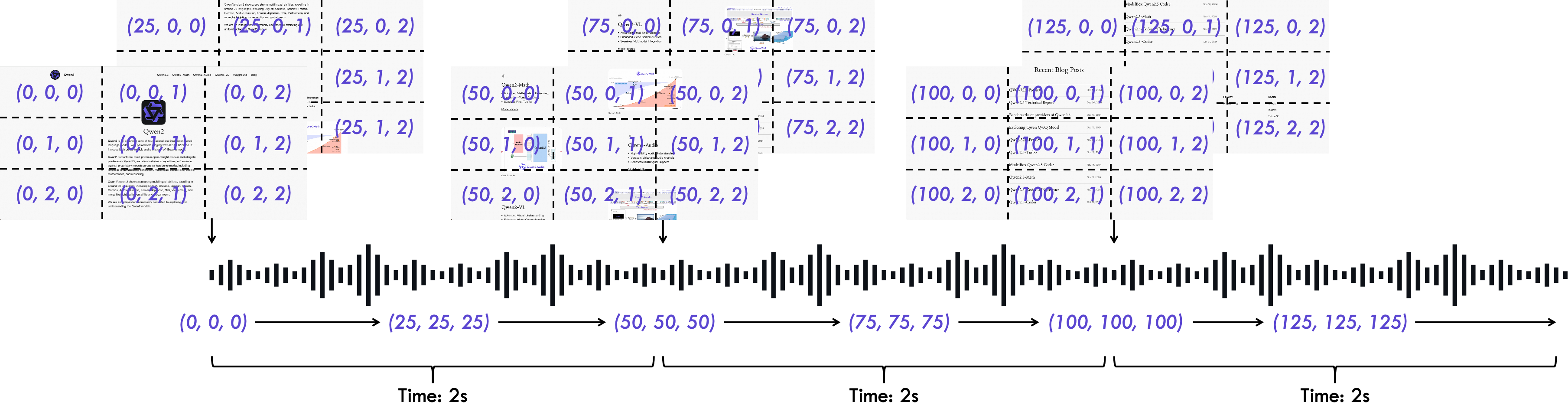}
    \caption{An illustration of Time-aligned Multimodal RoPE~(TMRoPE).}
    \label{fig:ATMRoPE}
\end{figure}
\paragraph{Text, Audio, Image and Video (w/o Audio).} Thinker processes text, audio, images, and video~(without the audio track) by converting them into a series of hidden representations for input. For tokenizing text, we use Qwen's tokenizer~\citep{qwen2}, which applies byte-level byte-pair encoding with a vocabulary comprising 151,643 regular tokens. Regarding audio input and audio from videos, we resample it to a frequency of 16kHz and transform the raw waveform into a 128-channel mel-spectrogram with a window size of 25ms and a hop size of 10ms. We adopt the audio encoder from Qwen2-Audio~\citep{qwen2-audio}, to make each frame of audio representation roughly corresponds to a 40ms segment of the original audio signal. Furthermore, we employ the vision encoder from Qwen2.5-VL~\citep{bai2025qwen25}, which is based on the Vision Transformer (ViT) model with approximately 675 million parameters, enabling it to effectively handle both image and video inputs. The vision encoder employs a mixed training regimen incorporating both image and video data, ensuring proficiency in image understanding and video comprehension. To preserve video information as completely as possible while adapting to the audio sampling rate, we sample the video using a dynamic frame rate. Additionally, for consistency, each image is treated as two identical frames.

\paragraph{Video and TMRoPE.} 
We propose a time-interleaving algorithm for audio and video, along with a novel position encoding approach. As shown in Figure~\ref{fig:ATMRoPE}, TMRoPE encodes the 3-D positional information of multimodal inputs, which is Multimodal Rotary Position Embedding (M-RoPE)~\citep{qwenvl} with absolute temporal positions. This is achieved by deconstructing the original rotary embedding into three components: temporal, height, and width. For text inputs, these components utilize identical position IDs, making M-RoPE functionally equivalent to 1D-RoPE. Similarly, for audio inputs, we also use identical position IDs and introduce absolute temporal position encoding, with one temporal ID corresponding to 40ms.

When processing images, the temporal IDs of each visual token remain constant, while distinct IDs are assigned to the height and width components based on the token’s position in the image. When the input is video with audio, the audio is still encoded with identical position IDs for every 40ms per frame, and the video is treated as a series of images with temporal ID increments for each frame, while the height and width components follow the same ID assignment pattern as images. Since the frame rate in video is not fixed, we dynamically adjust the temporal IDs between frames based on the actual time corresponding to each frame to ensure that one temporal ID corresponds to 40ms. In scenarios where the model’s input encompasses multiple modalities, position numbering for each modality is initialized by incrementing the maximum position ID of the preceding modality by one. TMRoPE enhances positional information modeling, maximizing the integration of various modalities, enabling \method to simultaneously understand and analyze information from multiple modalities.

After incorporating positional information into each modality, we arrange the representations in order. To enable the model to receive both visual and auditory information simultaneously, as shown in Figure~\ref{fig:ATMRoPE}, we have a special design for video with audio called the time-interleaving method, which segments the representation in the video with audio into chunks every 2 seconds according to the actual time. We then arrange the visual representation at the front and the audio representation at the back within the 2 seconds, interleaving the representations of the video with audio.

\subsection{Generation}
\paragraph{Text.} Text is generated directly by Thinker. The logic of text generation is fundamentally the same as that employed by widely used LLMs, which generate text through autoregressive sampling based on the probability distribution over the vocabulary. The generation process may incorporate techniques such as repetition penalty and top-p sampling to enhance its diversity.

\paragraph{Speech.} Talker receives both high-level representations and embeddings of the text tokens sampled by Thinker. The integration of high-dimensional representations and discrete sampling tokens is essential in this context. As a streaming algorithm, voice generation must anticipate the content's tone and attitude before the entire text is fully generated. The high-dimensional representations provided by Thinker implicitly convey this information, enabling a more natural streaming generation process. Furthermore, Thinker's representations primarily express semantic similarity in the representational space rather than phonetic similarity. Consequently, even phonetically distinct words may have very similar high-level representations, necessitating the input of sampled discrete tokens to eliminate such uncertainty.

We designed an efficient speech codec named \textit{qwen-tts-tokenizer}. \textit{qwen-tts-tokenizer} efficiently represents key information of speech and can be decoded to speech streamingly through a causal audio decoder. After receiving the information, Talker starts to autoregressively generate audio tokens and text tokens. The generation of speech does not require word-level and timestamp-level alignment with the text. This significantly simplifies the requirements for training data and the inference process.

\subsection{Designs for Streaming}
In the context of streaming audio and video interactions, the initial packet latency is a critical indicator of the system's streaming performance. This latency is influenced by several factors: 1) the delay caused by the processing of multimodal information inputs; 2) the latency from the moment the first text input is received until the first voice token is output; 3) the delay in converting the first segment of speech into audio; and 4) the inherent latency of the architecture itself, which is related to model size, computational FLOPs, and other factors. This paper will subsequently discuss the algorithmic and architectural improvements made to reduce these latencies across these four dimensions.

\paragraph{Support Prefilling.} Chunked-prefills is a mechanism widely used in modern inference framework. To support it in modalities interation, we modified the audio and visual encoders to support block-wise attention along the temporal dimension. Specifically, the audio encoder is changed from full attention over the entire audio to performing attention in blocks of 2 seconds each. The vision encoder utilizes flash attention for efficient training and inference with a simple MLP layer that merges adjacent 2×2 tokens into a single token. The patch size is set to 14, which allows images of different resolutions to be packed into a sequence.

\paragraph{Streaming Codec Generation.} To facilitate the streaming of audio, especially for extended sequences, we propose a sliding window block attention mechanism that restricts the current token's access to a limited context. Specifically, we utilize a Flow-Matching~\citep{flowmatching} DiT model. The input code is transformed into a mel-spectrogram using Flow-Matching, followed by a modified BigVGAN~\citep{sangbigvgan} to reconstruct the generated mel-spectrogram back into the waveform.

\begin{figure}[h]
    \centering
    \includegraphics[width=0.3\textwidth]{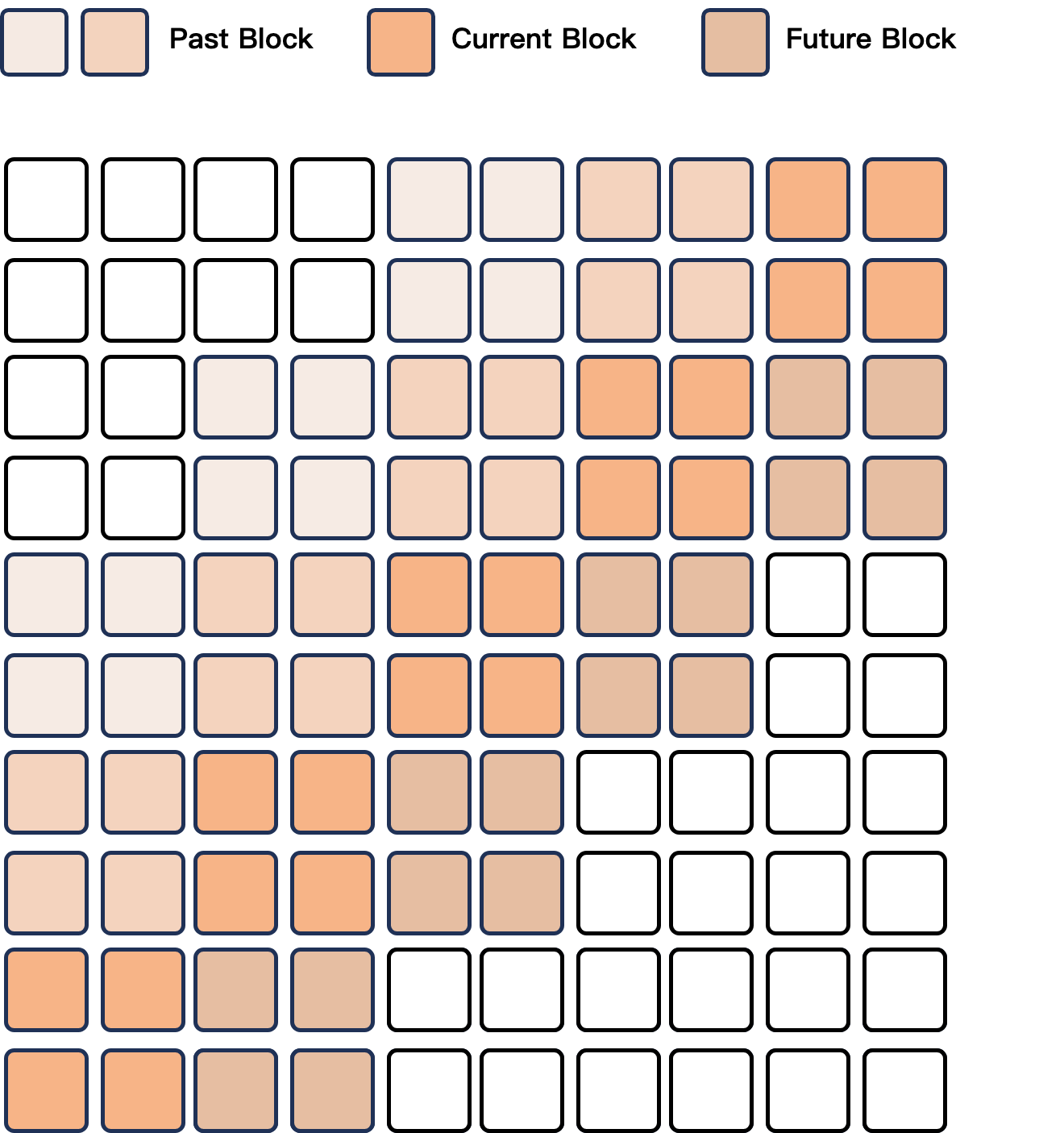}
    \caption{
An illustration of sliding window block attention mechanism in DiT for codec to wav generation.}
    \label{fig:attention_mask}
\end{figure}

As shown in Figure~\ref{fig:attention_mask}, to generate waveforms from code, we group adjacent codes into blocks and use these for our attention mask. We limit the DiT's receptive field to 4 blocks, including a lookback of 2 blocks and a lookahead of 1 block. During decoding, we generate the mel-spectrum in chunks using Flow Matching, ensuring that each code chunk has access to the necessary contextual blocks. This approach enhances the quality of streaming outputs by maintaining contextual information. We also use this chunk-by-chunk method for BigVGAN's fixed receptive field to facilitate streaming waveform generation




\section{Pre-training}
\method consists of three training stages. In the first stage, we lock the LLM parameters and focus exclusively on training the vision encoder and audio encoder, utilizing a vast corpus of audio-text and image-text pairs to enhance semantic understanding within the LLM. In the second stage, we unfreeze all parameters and train with a wider range of multimodal data for more comprehensive learning. In the final stage, we use data with a sequence length of 32k to enhance the model's ability to understand complex long-sequence data.

The model is pre-trained on a diverse dataset that includes various types such as image-text, video-text, video-audio, audio-text and text corpus. We replace the hierarchical tags with the natural language prompts following Qwen2-Audio~\citep{qwen2audio}, which can improve better generalization ability and better instruction following ability.

During the initial pre-training phase, the LLM component of \method is initialized using the parameters from Qwen2.5~\citep{qwen2.5}, while the vision encoder is the same as Qwen2.5-VL, and the audio encoder is initialized with the Whisper-large-v3~\citep{Whisper}. The two encoders are trained separately on the fixed LLM, with both initially focusing on training their respective adapters before training the encoders. This foundational training is crucial in equipping the model with a robust understanding of core visual-textual and audio-textual correlations and alignments.

The second phase of pre-training marks a significant advancement by incorporating an additional 800 billion tokens of image and video related data, 300 billion tokens of audio related data, and 100 billion tokens of video with audio related data. This phase introduces a larger volume of mixed multimodal data and a wider variety of tasks, which enhances the interaction and deepens the understanding between auditory, visual, and textual information. The inclusion of multimodal, multitask datasets is crucial for developing the model's ability to handle multiple tasks and modalities simultaneously, a vital capability for managing complex real-world datasets. Moreover, pure text data plays an essential role in maintaining and improving language proficiency.

To improve training efficiency, we limited the maximum token length to 8192 tokens in the previous stages. Then, we incorporat long audio and long video data and extend the original text, audio, image, and video  data to 32,768 tokens for training. Experimental results indicate that our data shows significant improvement in supporting long sequence data

\section{Post-training}\label{sec:post}
\subsection{Data Format}
\begin{tcolorbox}[colback=black!5!white,colframe=black!75!black,title=The Dataset Format Example of ChatML]
\textcolor{blue}{<|im\_start|>}user

<|vision\_start|>Video.mp4 [Two people are talking in the video]<|vision\_end|>What are the people in the video saying?\textcolor{blue}{<|im\_end|>}

\textcolor{blue}{<|im\_start|>}assistant

\textcolor{blue}{Both pictures are of SpongeBob SquarePants.The person in the red clothes says, "Hello, how's the weather today?" The person in the black clothes responds, "Hello, the weather is quite nice today."<|im\_end|>}

\textcolor{blue}{<|im\_start|>}user

<|vision\_start|>Video.mp4 [A person in the video is saying, "Please describe the person in front of you."]<|vision\_end|>\textcolor{blue}{<|im\_end|>}

\textcolor{blue}{<|im\_start|>}assistant

\textcolor{blue}{The person in front of you is wearing glasses and a brown jacket over a blue shirt. They appear to be speaking or reacting to something, as their mouth is open and they seem engaged. The background shows a room with a wall-mounted air conditioner, a clothing rack with various garments hanging on it, and a large screen displaying an image of a cityscape at night. The lighting in the room is warm and cozy.<|im\_end|>}
\end{tcolorbox}

\subsection{Thinker}
During the post-training phase, we employ instruction-following data with ChatML~\citep{chatml} format for instruction-finetuning. Our dataset incorporates pure text-based dialogue data, visual-modality conversation data, audio-modality conversation data and mix-modality conversation data. 

\subsection{Talker}
We introduced a three-stage training process for Talker, allowing \method to generate text and speech responses simultaneously. In the first stage, we train Talker to learn context continuation. The second stage utilized DPO~\citep{rafailov2024direct} to enhance the stability of speech generation. In the third stage, we applied multi-speaker instruction fine-tuning to improve the naturalness and controllability of the speech responses.

During the In-Context Learning (ICL) training phase, in addition to utilizing text supervision similar to that of Thinker, we perform a speech continuation task through next-token prediction, leveraging an extensive dataset of dialogues that incorporate multimodal contexts and spoken responses. Talker learns to establish a monotonic mapping from semantic representation to speech, while also acquiring the ability to express speech with diverse attributes that are contextually appropriate, such as prosody, emotion, and accent. Additionally, we implement timbre disentanglement techniques to prevent the model from associating specific voices with infrequent textual patterns.

\begin{equation}
    \mathcal{L}_{\text{DPO}}(\mathcal{P}_\theta; \mathcal{P}_{\text{ref}}) = -\mathbb{E}_{(\bm{x}, \bm{y_w}, \bm{y_l}) \sim \mathcal{D}} \left[ \log \sigma \left( \beta \log \frac{\mathcal{P}_\theta(\bm{y_w} \mid \bm{x})}{\mathcal{P}_{\text{ref}}(\bm{y_w} \mid \bm{x})} - \beta \log \frac{\mathcal{P}_\theta(\bm{y_l} \mid \bm{x})}{\mathcal{P}_{\text{ref}}(\bm{y_l} \mid \bm{x})} \right) \right].
\end{equation}

To broaden the coverage of speakers and scenarios, the pretraining data inevitably contains label noise and pronunciation errors, leading to model hallucinations. 
To mitigate this issue, we introduce a reinforcement learning phase to improve the stability of speech generation.  Specifically, for each request and response text paired with the reference speech, we build a dataset $\mathcal{D}$ with the triplet data $(\bm{x}, \bm{y_w}, \bm{y_l})$, where $\bm{x}$ is the input sequence with input text, and $\bm{y_w}$ and $\bm{y_l}$ are the good and bad generated speech sequences respectively. We rank these samples based on their reward scores associated with word error rate~(WER) and the punctuation pause error rate.

Lastly, we performed speaker fine-tuning on the aforementioned base model, enabling Talker to adopt specific voices and improve its naturalness. 

\section{Evaluation}
\label{sec:experiment}
We conduct comprehensive evaluation of \method. The model is divided into two main categories: understanding~(X$\to$Text) and speech generation~(X$\to$Speech). 

\subsection{Evaluation of X$\to$Text}
In this section, we evaluate \method's ability to comprehend various multimodal inputs (text, audio, image, and video) and generate textual responses.

\paragraph{Text$\to$Text} Our evaluation of \method on text $\to$ text primarily focuses on general evaluation, mathematics \& science ability and coding ability. Specifically, we utilize MMLU-Pro~\citep{mmlupro}, MMLU-redux~\citep{mmluredux} and Livebench0803~\citep{livebench} for general evaluation, GPQA~\citep{gpqa}, GSM8K~\citep{gsm8k} and MATH~\citep{math} for mathematics \& science, HumanEval~\citep{humaneval}, MBPP~\citep{mbpp}, MultiPL-E~\citep{multiple} and LiveCodeBench 2305-2409~\citep{livecodebench} for coding.

\paragraph{Audio$\to$Text} The evaluation of \method for audio $\to$ text includes audio understanding, audio reasoning, and voice-chatting. Specifically, we perform a comprehensive evaluation on Automatic Speech Recognition (ASR), Speech-to-Text Translation (S2TT), Speech Entity Recognition (SER), Vocal Sound classification (VSC) and Music, which assesses the performance of \method on a broad range of audio understanding tasks. We utilize MMAU~\citep{sakshi2024mmaumassivemultitaskaudio} for audio reasoning tasks, VoiceBench~\citep{chen2024voicebench} and a self-curated speech-instruction benchmark for voice-chatting tasks.

\paragraph{Image$\to$Text} The evaluation of \method for image $\to$ text primarily emphasizes the performance
in college-level problems, math, general visual question answering and OCR-related tasks. Specifically, we utilize MMMU~\citep{yue2023mmmu} and MMMU-Pro~\citep{mmmupro} for college-level problems evaluation, MathVista~\citep{mathvista} and MathVision~\citep{mathvision} for math. For general visual question answering, we evaluate the performance on benchmark datasets such as MMBench-V1.1~\citep{MMBench}, MMVet~\citep{yu2024mm}, MMStar~\citep{chen2024we}, MME~\citep{fu2023mme}, MuirBench~\citep{wang2024muirbenchcomprehensivebenchmarkrobust}, CRPE~\citep{wang2024allseeingprojectv2general}, RealWorldQA~\citep{X.AI.}, 
MMERealWorld~\citep{mme-realworld}, and MM-MT-Bench~\citep{agrawal2024pixtral12b}. 
Additionally, we evaluate \method on various OCR benchmarks, such as AI2D~\citep{kembhavi2016diagram}, TextVQA~\citep{textvqa}, DocVQA~\citep{docvqa}, ChartQA~\citep{masry2022chartqa}, and OCRBench\_v2~\citep{fu2024ocrbenchv2improvedbenchmark}. Furthermore, we also evaluate the visual grounding capability of our model on the referring expression comprehension benchmarks~\citep{kazemzadeh-etal-2014-referitgame, mao2016generationcomprehensionunambiguousobject}, object detection in the wild~\citep{li2022groundedlanguageimagepretraining} and a self-curated point grounding benchmark.

\paragraph{Video~(w/o Audio)$\to$Text} We assess our model on several representative video understanding tasks like Video-MME~\citep{fu2024video},
MVBench~\citep{li2024mvbench}, and EgoSchema~\citep{mangalam2023egoschema}.

\paragraph{Multimodality$\to$Text}
We demonstrate the ability of our model for mixed-modality (image, audio and text) prompts on OmniBench~\citep{li2024omnibench}.

\subsubsection{Performance of Text$\to$Text}
We compare \method with other leading large language model of similar size (7B). As shown in Table \ref{tab:pure-text}, the performance of \method generally falls between Qwen2-7B and Qwen2.5-7B. Our model outperforms Qwen2-7B on most benchmarks, such as MMLU-Pro, MMLU-redux, MATH, GSM8K, MBPP, MultiPL-E and LiveCodeBench, which demonstrates the exceptional capabilities of our model for Text$\to$Text.

\begin{table}[htbp]
\centering

\caption{\textbf{Text $\to$ Text performance of 7B+ pure text models and \method}}
\label{tab:pure-text}
\small
\setlength{\tabcolsep}{2.6pt}
\begin{tabular}{@{}lccccc@{}}
\toprule
\textbf{Datasets} & \textbf{Gemma2-9B} & \textbf{Llama3.1-8B} & \textbf{Qwen2-7B} & \textbf{Qwen2.5-7B} & \textbf{\method-7B} \\
\midrule
\multicolumn{6}{c}{\textit{General Tasks}} \\
\midrule
MMLU-Pro & 52.1 & 48.3 & 44.1 & \textbf{56.3} & 47.0 \\
MMLU-redux & 72.8 & 67.2 & 67.3 & \textbf{75.4} & 71.0 \\
LiveBench$_{\text{0831}}$ & 30.6 & 26.7 & 29.2 &\textbf{35.9} & 29.6 \\
\midrule
\multicolumn{6}{c}{\textit{Mathematics \& Science Tasks}} \\
\midrule
GPQA & 32.8 & 32.8 & 34.3 &\textbf{36.4} & 30.8 \\
MATH & 44.3 & 51.9 & 52.9 &\textbf{75.5} & 71.5 \\
GSM8K & 76.7 & 84.5 & 85.7 &\textbf{91.6} & 88.7 \\
\midrule
\multicolumn{6}{c}{\textit{Coding Tasks}} \\
\midrule
HumanEval & 68.9 & 72.6 &79.9 &\textbf{84.8} &78.7 \\
MBPP & 74.9 & 69.6 & 67.2 &\textbf{79.2} &73.2 \\
MultiPL-E & 53.4 & 50.7 & 59.1 &\textbf{70.4} &65.8 \\
LiveCodeBench$_{\text{2305-2409}}$ & 18.9 & 8.3 &23.9 &\textbf{28.7} & 24.6\\
\bottomrule
\end{tabular}
\end{table}

\subsubsection{Performance of Audio$\to$Text}
We compare \method with other leading specialist or generalist models on diverse audio understanding, audio reasoning, and voice-chatting benchmarks. As shown in Table \ref{tab:audio_analysis_table} and \ref{tab:audio_analysis_table_1}, \method delivers better or comparable performance with other state-of-the-art methods on audio understanding. For instance, it achieves superior ASR and S2TT performance on Fleurs\_zh, CommonVoice\_en, CommonVoice\_zh, CoVoST2\_en-de and CoVoST2\_zh-en test sets, surpassing previous state-of-the-art models like Whisper-large-v3, Qwen2Audio, MinMo and other Omni models. \method also achieves state-of-the-art performance on general audio understanding tasks like music and VSC. Additionally, \method achieves state-of-the-art results on audio reasoning with superior performance on sound, music and speech subsets of MMAU benchmark. These results demonstrate the powerful capabilities of \method in general audio understanding and reasoning.

Additionally, on VoiceBench, \method achieves an impressive average score of 74.12, surpassing other audio language models and omni models of similar size. This showcases our model's strong capabilities in speech interaction.
To further explore the performance of diverse speech interaction, we convert text instructions from several pure-text benchmarks into speech and evaluate \method, Qwen2-Audio and Qwen2-7B on the in-house voice-chat benchmark. About 90\% of text-instructions are utilized. We use speech instruction for \method and Qwen2-Audio, and text instruction for Qwen2-7B. 
As shown in Table \ref{tab:voicechat-in-house}, compared to Qwen2-Audio, \method significantly narrowes the gap with Qwen2-7B, which uses text instructions. This reflects our model's substantial progress in diversified end-to-end speech interaction.
\begin{table}[htbp]
\centering
\caption{\textbf{Audio $\to$ text performance of State-of-the-art and \method}}
\vspace{-2mm}
\resizebox{0.8\textwidth}{!}{
\begin{tabular}{cccc}
\toprule
\textbf{Datasets} & \textbf{Model} & \textbf{Performance}
\\ 
\midrule \multicolumn{3}{c}{\textit{ASR}} \\ 
\midrule 
\multirow{11}{*}{\begin{tabular}[c]{@{}c@{}}\textbf{Librispeech}\\ \textit{dev-clean} | \textit{dev-other} | \\ \textit{test-clean} | \textit{test-other} \end{tabular}}      
   & SALMONN~\citep{anonymous2023salmonn}  & - | - | 2.1 | 4.9  
\\ & SpeechVerse~\citep{das2024speechverse}  & - | - | 2.1 | 4.4   
\\ & Whisper-large-v3~\citep{Whisper} & - | - | 1.8 | 3.6
\\ & Llama-3-8B~\citep{dubey2024llama} & - | - | - | 3.4
\\ & Llama-3-70B~\citep{dubey2024llama} & - | - | - | 3.1
\\ & Seed-ASR-Multilingual~\citep{bai2024seed} & - | - | \textbf{1.6} | \textbf{2.8}
\\ & MiniCPM-o~\citep{yao2024minicpm} & - | - | 1.7 | -
\\ & MinMo~\citep{chen2025minmomultimodallargelanguage} & - | - | 1.7 | 3.9
\\ & Qwen-Audio~\citep{chu2023qwen}  & 1.8 | 4.0 | 2.0 | 4.2   
\\ & Qwen2-Audio~\citep{qwen2audio} & \textbf{1.3} | \textbf{3.4} | \textbf{1.6} | 3.6 
\\ & \method-7B & 1.6 | 3.5 | 1.8 | 3.4
\\ 
\midrule  \multirow{3}{*}{\begin{tabular}[c]{@{}c@{}}\textbf{Common Voice 15} \\ \textit{en} | \textit{zh} | \textit{yue} | \textit{fr} \end{tabular}} 
   & Whisper-large-v3~\citep{Whisper} & 9.3 | 12.8 | 10.9 | 10.8  
\\ & MinMo~\citep{chen2025minmomultimodallargelanguage} & 7.9 | 6.3 | 6.4 | 8.5
\\ & Qwen2-Audio~\citep{qwen2audio} & 8.6 | 6.9 | \textbf{5.9} | 9.6 
\\ & \method-7B & \textbf{7.6} | \textbf{5.2} | 7.3 | \textbf{7.5}

\\ 
\midrule \multirow{5}{*}{\begin{tabular}[c]{@{}c@{}}\textbf{Fleurs} \\ \textit{zh} | \textit{en} \end{tabular}}        
   & Whisper-large-v3~\citep{Whisper}  & 7.7 | 4.1
\\ & Seed-ASR-Multilingual~\citep{bai2024seed} & - | \textbf{3.4}
\\ & Megrez-3B-Omni~\citep{Megrez-3B-Omni} & 10.8 | -
\\ & MiniCPM-o~\citep{yao2024minicpm} & 4.4 | -
\\ & MinMo~\citep{chen2025minmomultimodallargelanguage} & \textbf{3.0} | 3.8
\\ & Qwen2-Audio~\citep{qwen2audio} & 7.5 | - 
\\ & \method-7B & \textbf{3.0} | 4.1

\\ 
\midrule  \multirow{3}{*}{\begin{tabular}[c]{@{}c@{}}\textbf{Wenetspeech} \\ \textit{test-net} | \textit{test-meeting} \end{tabular}}        
   & Seed-ASR-Chinese~\citep{bai2024seed} & \textbf{4.7} | \textbf{5.7}
\\ & Megrez-3B-Omni~\citep{Megrez-3B-Omni} & - | 16.4
\\ & MiniCPM-o~\citep{yao2024minicpm} & 6.9 | -
\\ & MinMo~\citep{chen2025minmomultimodallargelanguage} & 6.8 | 7.4
\\ & \method-7B & 5.9 | 7.7


\\ 
\midrule  \multirow{3}{*}{\textbf{Voxpopuli-V1.0-en}}      
   & Llama-3-8B~\citep{dubey2024llama} & 6.2 
\\ & Llama-3-70B~\citep{dubey2024llama} & \textbf{5.7}
\\ & \method-7B & 5.8


\\ \midrule \multicolumn{3}{c}{\textit{S2TT}} 
\\\midrule   \multirow{5}{*}{\begin{tabular}[c]{@{}c@{}}\textbf{CoVoST2} \\ \textit{en-de} | \textit{de-en} | \\ \textit{en-zh} | \textit{zh-en} \end{tabular}} 
    &SALMONN~\citep{anonymous2023salmonn}    & 18.6 | - | 33.1 | -         
\\  & SpeechLLaMA~\citep{speechllama} & - | 27.1 | - | 12.3   
\\  & BLSP~\citep{wang2023blsp}  & 14.1 | - | - | -    
\\  & MiniCPM-o~\citep{yao2024minicpm} & - | - | \textbf{48.2} | 27.2
\\  & MinMo~\citep{chen2025minmomultimodallargelanguage} & - | \textbf{39.9} | 46.7 | 26.0
\\  & Qwen-Audio~\citep{chu2023qwen}  & 25.1 | 33.9 | 41.5 | 15.7   
\\  & Qwen2-Audio~\citep{qwen2audio}  & 29.9 | 35.2 | 45.2 | 24.4
\\  & \method-7B &\textbf{30.2} | 37.7 | 41.4 | \textbf{29.4}




\\       
\bottomrule

\end{tabular}}
\label{tab:audio_analysis_table}
\end{table}

\begin{table}[t!]
\centering
\caption{\textbf{Audio $\to$ text performance of State-of-the-art and \method}}
\vspace{-2mm}
\resizebox{\textwidth}{!}{
\begin{tabular}{cccc}
\toprule
\textbf{Datasets} & \textbf{Model} & \textbf{Performance}
\\ \midrule \multicolumn{3}{c}{\textit{SER}} 
\\ \midrule \multirow{7}{*}{\textbf{Meld}} 
   & WavLM-large~\citep{wavlm}  & 0.542
\\ & MiniCPM-o~\citep{yao2024minicpm} & 0.524
\\ & Qwen-Audio~\citep{chu2023qwen} & 0.557
\\ & Qwen2-Audio~\citep{qwen2audio} & 0.553  
\\ & \method-7B & \textbf{0.570}


\\ \midrule \multicolumn{3}{c}{\textit{VSC}} 
\\ \midrule  \multirow{5}{*}{\textbf{VocalSound}}     
   & CLAP~\citep{CLAP} & 0.495                
\\ & Pengi~\citep{Pengi} & 0.604                
\\ & Qwen-Audio~\citep{chu2023qwen} & 0.929                 
\\ & Qwen2-Audio~\citep{qwen2audio} & \textbf{0.939} 
\\ & \method-7B & \textbf{0.939}


\\ \midrule \multicolumn{3}{c}{\textit{Music}} 
\\ \midrule \multirow{2}{*}{\begin{tabular}[c]{@{}c@{}}\textbf{GiantSteps} \\ \textit{Tempo} \end{tabular}}  
 & LLark-7B~\citep{gardner2023llark} & 0.86
\\ & \method-7B & \textbf{0.88}
\\
\midrule  \multirow{2}{*}{\textbf{MusicCaps}} 
   & LP-MusicCaps~\citep{doh2023lp} & 0.291 | 0.149 | 0.089 | \textbf{0.061} | \textbf{0.129} | 0.130 
\\ & \method-7B & \textbf{0.328}  | \textbf{0.162} | \textbf{0.090} | 0.055 | 0.127 | \textbf{0.225} 
 
\\ \midrule \multicolumn{3}{c}{\textit{Audio Reasoning}} 
\\ \midrule  \multirow{3}{*}{\begin{tabular}[c]{@{}c@{}}\textbf{MMAU} \\ \textit{Sound} | \textit{Music} | \\ \textit{Speech} | \textit{Avg} \end{tabular}} 
   & Gemini-Pro-V1.5~\citep{team2024gemini} & 56.75 | 49.40 | 58.55 |  54.90
\\ & Qwen2-Audio~\citep{qwen2audio} & 54.95 | 50.98 | 42.04 | 49.20
\\ & \method-7B & \textbf{67.87} | \textbf{69.16} | \textbf{59.76} | \textbf{65.60}

\\ \midrule \multicolumn{3}{c}{\textit{Voice Chatting}} 
\\\midrule \multirow{7}{*}{\begin{tabular}[c]{@{}c@{}}\textbf{VoiceBench} \\ \textit{AlpacaEval} | \textit{CommonEval} | \\ \textit{SD-QA} | \textit{MMSU} \end{tabular}}   
   & Ultravox-v0.4.1-LLaMA-3.1-8B   & \textbf{4.55} | 3.90 | 53.35 | 47.17
\\ & MERaLiON~\citep{he2024meralion}  & 4.50 | 3.77 | 55.06 | 34.95    
\\ & Megrez-3B-Omni~\citep{Megrez-3B-Omni} & 3.50 | 2.95 | 25.95 | 27.03
\\ & Lyra-Base~\citep{zhong2024lyra} & 3.85 | 3.50 | 38.25 | 49.74   
\\ & MiniCPM-o~\citep{yao2024minicpm} & 4.42 | \textbf{4.15} | 50.72 | 54.78
\\ & Baichuan-Omni-1.5~\citep{li2025baichuan} & 4.50 | 4.05 | 43.40 | 57.25
\\ & Qwen2-Audio~\citep{qwen2audio} & 3.74 | 3.43 | 35.71 | 35.72   
\\ & \method-7B & 4.49 | 3.93 | \textbf{55.71} | \textbf{61.32}
\\ \midrule \multirow{7}{*}{\begin{tabular}[c]{@{}c@{}}\textbf{VoiceBench} \\ \textit{OpenBookQA} | \textit{IFEval} | \\ \textit{AdvBench} | \textit{Avg}  \end{tabular}}
   & Ultravox-v0.4.1-LLaMA-3.1-8B   & 65.27 | \textbf{66.88} | 98.46 | 71.45
\\ & MERaLiON~\citep{he2024meralion}  & 27.23 | 62.93 | 94.81 | 62.91    
\\ & Megrez-3B-Omni~\citep{Megrez-3B-Omni} & 28.35 | 25.71 | 87.69 | 46.25
\\ & Lyra-Base~\citep{zhong2024lyra} & 72.75 | 36.28 | 59.62 | 57.66
\\ & MiniCPM-o~\citep{yao2024minicpm} & 78.02 | 49.25 | 97.69 | 71.69
\\ & Baichuan-Omni-1.5~\citep{li2025baichuan} & 74.51 | 54.54 | 97.31 | 71.14
\\ & Qwen2-Audio~\citep{qwen2audio} & 49.45 | 26.33 | 96.73 | 55.35  
\\ & \method-7B & \textbf{81.10} | 52.87 | \textbf{99.42} | \textbf{74.12}
\\
\bottomrule

\end{tabular}}
\label{tab:audio_analysis_table_1}
\end{table}

\begin{table}[t]
\centering
\caption{Performance of \method and other models for Chatting, $^{\ast}$ means that approximately 90\% of text instructions suitable for speech are used.}
\label{tab:voicechat-in-house}
\small
\setlength{\tabcolsep}{2.6pt}
\begin{tabular}{@{}lccc@{}}
\toprule
\textbf{Datasets} & \textbf{Qwen2-7B} (text) & \textbf{Qwen2-Audio} & \textbf{\method-7B} \\
\midrule
MMLU$^{\ast}$ & 69.3 & 33.2 & 65.6 \\
CEval$^{\ast}$ & 78.4 & 38.6 & 61.1 \\
IFEval$^{\ast}$ & 53.3 & 15.6 & 41.7 \\
GSM8K$^{\ast}$  & 82.3 & 18.4 & 85.4 \\
Math23K$^{\ast}$ & 92.3 & 23.0 & 87.1 \\
Math401$^{\ast}$ & 75.5 & 20.4 & 62.2 \\
\bottomrule
\end{tabular}
\end{table}

\subsubsection{Performance of Image $\to$ Text}
To comprehensively evaluate the capabilities on Image $\to$ Text, we compare \method with the recent state-of-the-art large vision language model Qwen2.5-VL-7B and other best-performing omni models. As illustrated in Table \ref{tab:image-benchmark}, \method demonstrates comparable performance to Qwen2.5-VL-7B, and attains better results on MMMU, MathVision, MMBench-V1.1-EN, TextVQA, DocVQA and ChartQA than any other open-sourced omni models.  Additionally, \method also surpasses GPT-4o-mini on most benchmarks. These results reveal the excellent capability of our model on image understanding.

\begin{table}[t]
\centering
\caption{\textbf{Image $\to$ Text performance of 7B+ models and \method}}
\label{tab:image-benchmark}
\small
\setlength{\tabcolsep}{2.6pt}
\begin{tabular}{@{}lcccc@{}}
\toprule
\textbf{Datasets} & \textbf{GPT-4o-mini} & \textbf{Qwen2.5-VL-7B} & \textbf{Other Best} &\textbf{\method-7B} \\
\midrule
\multicolumn{5}{c}{\textit{College-level Problems}} \\
\midrule
MMMU\textsubscript{val} & \textbf{60.0} & 58.6 & 53.9~\citep{li2025baichuan} & 59.2 \\
MMMU-Pro\textsubscript{overall} & 37.6 & \textbf{38.3} & - & 36.6 \\
\midrule
\multicolumn{5}{c}{\textit{Mathematical}} \\
\midrule
MathVista\textsubscript{testmini} & 52.5 & 68.2 & \textbf{71.9}~\citep{yao2024minicpm} & 67.9 \\
MathVision\textsubscript{full} & - & \textbf{25.1} & 23.1~\citep{yao2024minicpm}  & 25.0 \\
\midrule
\multicolumn{5}{c}{\textit{General Visual Question Answering}} \\
\midrule
MMBench-V1.1-EN\textsubscript{test} & 76.0 & \textbf{82.6} & 80.5~\citep{yao2024minicpm} & 81.8 \\
MMVet\textsubscript{turbo} & 66.9 & 67.1 & \textbf{67.5}~\citep{yao2024minicpm} & 66.8 \\
MMStar & 54.8 & 63.9 & \textbf{64.0}~\citep{yao2024minicpm} & \textbf{64.0} \\
MME\textsubscript{sum} & 2003 & 2347 & \textbf{2372}~\citep{yao2024minicpm} & 2340 \\
MuirBench & - & \textbf{59.6} & - & 59.2 \\
CRPE\textsubscript{relation} & - &76.4 & - & \textbf{76.5} \\
RealWorldQA\textsubscript{avg} & - &68.5 & \textbf{71.9}~\citep{Megrez-3B-Omni} & 70.3 \\
MME-RealWorld\textsubscript{en} & - &57.4 & - &\textbf{61.6} \\
MM-MT-Bench & - & \textbf{6.3} & - & 6.0 \\
\midrule
\multicolumn{5}{c}{\textit{OCR-related Tasks}} \\
\midrule
AI2D & - & 83.9 & \textbf{85.8}~\citep{yao2024minicpm} & 83.2\\
TextVQA\textsubscript{val} & - & \textbf{84.9} & 83.2~\citep{li2025baichuan} & 84.4 \\
DocVQA\textsubscript{test} & - & \textbf{95.7} & 93.5~\citep{yao2024minicpm} & 95.2 \\
ChartQA\textsubscript{test Avg} & - & \textbf{87.3} &84.9~\citep{li2025baichuan}  &85.3 \\
OCRBench\_V2\textsubscript{en} & - &56.3 & - & \textbf{57.8} \\
\bottomrule
\end{tabular}
\end{table}

For visual grounding, we compare \method with Qwen2.5-VL-7B and other leading LVLMs including Gemini and Grounding-DINO~\citep{liu2024groundingdinomarryingdino}. As illustrated in Table \ref{tab:image-grounding}, our model outperforms other models across most benchmarks from box-grounding to point-grounding and achieves a good performance of 42.2mAP on open-vocabulary object detection, which reveals the strong visual grounding capability of our model.

\begin{table}[t]
\centering
\caption{\textbf{Grounding performance of \method and other models}}
\label{tab:image-grounding}
\small
\setlength{\tabcolsep}{2.6pt}
\begin{tabular}{@{}lcccc@{}}
\toprule
\textbf{Datasets} & \textbf{Gemini 1.5 Pro} & \textbf{Grounding DINO} & \textbf{Qwen2.5-VL-7B} &\textbf{\method-7B} \\
\midrule
Refcoco\textsubscript{val} & 73.2 & \textbf{90.6} & 90.0 & 90.5 \\
Refcoco\textsubscript{textA} & 72.9 & 93.2 & 92.5 & \textbf{93.5} \\
Refcoco\textsubscript{textB} & 74.6 & \textbf{88.2} & 85.4 & 86.6 \\
Refcoco+\textsubscript{val} & 62.5 & \textbf{88.2} & 84.2 & 85.4 \\
Refcoco+\textsubscript{textA} &63.9 &89.0 & 89.1 & \textbf{91.0} \\
Refcoco+\textsubscript{textB} &65.0 &75.9 & 76.9 & \textbf{79.3} \\
Refcocog\textsubscript{val} &75.2 & 86.1 &87.2 & \textbf{87.4} \\
Refcocog\textsubscript{test} & 76.2 & 87.0 & 87.2 & \textbf{87.9} \\
ODinW & 36.7 & \textbf{55.0} & 37.3 & 42.2 \\
PointGrounding & - & - & \textbf{67.3} & 66.5 \\
\bottomrule
\end{tabular}
\end{table}

\subsubsection{Performance of Video $\to$ Text}
Similar to Image$\to$Text, we compare \method with Qwen2.5-VL-7B and other omni models. As shown in Table \ref{tab:video-benchmark}, \method outperforms all other state-of-the-art open-sourced omni models and GPT-4o-Mini, and attains better or competitive results compared to Qwen2.5-VL-7B, which demonstrates the superior performance on video understanding.

\begin{table}[t]
\centering
\caption{\textbf{Video $\to$ text performance of 7B+ models and \method}}
\label{tab:video-benchmark}
\small
\setlength{\tabcolsep}{2.6pt}
\begin{tabular}{@{}lcccc@{}}
\toprule
\textbf{Datasets} & \textbf{GPT-4o-mini} & \textbf{Qwen2.5-VL-7B} & \textbf{Other Best} &\textbf{\method-7B} \\
\midrule
\multicolumn{5}{c}{\textit{Video Understanding}} \\
\midrule
Video-MME\textsubscript{w/o sub} & 64.8 & \textbf{65.1} & 63.9~\citep{yao2024minicpm} & 64.3 \\
Video-MME\textsubscript{w sub} & - & 71.6 & 67.9~\citep{yao2024minicpm} & \textbf{72.4} \\
MVBench & - & 69.6 & 67.2~\citep{zhong2024lyra} &\textbf{70.3} \\
EgoSchema\textsubscript{test} & - &65.0 & 63.2~\citep{zhong2024lyra} &\textbf{68.6} \\
\bottomrule
\end{tabular}
\end{table}

\subsubsection{Performance of Multimodality$\to$Text}
As shown in Table \ref{tab:omnibench}, \method achieves state-of-the-art performance on OmniBench, surpassing other Omni models by a large margin, which demonstrates the superiority of our model in multimodality understanding.

\begin{table}[t!]
\centering
\caption{Multimodality $\to$ Text performance of State-of-the-art and \method}
\label{tab:omnibench}
\vspace{-2mm}
\resizebox{\textwidth}{!}{
\begin{tabular}{cccc}
\toprule
\textbf{Datasets} & \textbf{Model} & \textbf{Performance}  
\\ \midrule \multicolumn{3}{c}{\textit{Multimodal Understanding}} 
\\\midrule \multirow{8}{*}{\begin{tabular}[c]{@{}c@{}}\textbf{OmniBench} \\ \textit{Speech} | \textit{Sound Event} | \\ \textit{Music} | \textit{Avg} \end{tabular}}   
   & Gemini-1.5-Pro~\citep{team2024gemini}  & 42.67\% | 42.26\% | 46.23\% | 42.91\% 
\\ &MIO-Instruct~\citep{wang2024mio} (7B) & 36.96\% | 33.58\% | 11.32\% | 33.80\%
\\ &AnyGPT (7B)~\citep{zhan2024anygpt}  & 17.77\% | 20.75\% | 13.21\% | 18.04\%
\\ &video-SALMONN (13B)~\citep{sun2024video} &34.11\% | 31.70\% | \textbf{56.60}\% | 35.64\%
\\ &UnifiedIO2-xlarge (3.2B)~\citep{lu2024unified} &39.56\% | 36.98\% | 29.25\% | 38.00\%
\\ &UnifiedIO2-xxlarge (6.8B)~\citep{lu2024unified} &34.24\% | 36.98\% | 24.53\% | 33.98\%
\\ &MiniCPM-o~\citep{yao2024minicpm} & - | - | - | 40.5\%
\\ &Baichuan-Omni-1.5~\citep{li2025baichuan} & - | - | - |  42.9\%
\\ &\method-7B & \textbf{55.25}\% | \textbf{60.00}\% | 52.83\% | \textbf{56.13}\%
\\
\bottomrule

\end{tabular}}
\label{tab:omnibench}
\end{table}

\subsection{Evaluation of X$\to$Speech}
In this section, we evaluate the speech generation capabilities of \method. Due to the lack of relevant assessments, the evaluation of speech generation focuses primarily speech generation given texts, similarity to text-to-speech (TTS), on two aspects: Zero-shot and Single-Speaker speech generation capabilities.

\begin{itemize}
    \item \textbf{Zero-Shot Speech Generation} We assessed the content consistency (WER) and speaker similarity (SIM) of our model in zero-shot speech generation on SEED~\citep{seedtts}.
    \item \textbf{Single-Speaker Speech Generation} We assessed the stability of our speaker fine-tuned model on the SEED~\citep{seedtts}, and evaluated the subjective naturalness (NMOS) of the generated speech on a self-created dataset.
\end{itemize}

\begin{table}[t!]
\centering
\caption{Zero-Shot Speech Generation}
\setlength{\tabcolsep}{2.6pt}
\begin{tabular}{@{}cll@{}}
\toprule
\textbf{Datasets} & \textbf{Model} & \textbf{Performance} \\
\midrule
\multicolumn{3}{c}{\textit{Content Consistency}} \\
\midrule 
\multirow{9}{*}{\begin{tabular}[c]{@{}c@{}}\textbf{SEED} \\ \textit{test-zh} | \textit{test-en} | \\ \textit{test-hard} \end{tabular}}   
   & Seed-TTS\textsubscript{ICL}~\citep{seedtts} & 1.11 | 2.24 | 7.58 \\ 
   & Seed-TTS\textsubscript{RL}~\citep{seedtts}  & \textbf{1.00} | 1.94 | \textbf{6.42} \\ 
   & MaskGCT~\citep{maskgct}                     & 2.27 | 2.62 | 10.27 \\ 
   & E2 TTS~\citep{e2tts}                        & 1.97 | 2.19 | - \\ 
   & F5-TTS~\citep{f5tts}                        & 1.56 | \textbf{1.83} | 8.67 \\ 
   & CosyVoice 2~\citep{cosyvoice2}              & 1.45 | 2.57 | 6.83 \\
   & CosyVoice 2-S~\citep{cosyvoice2}            & 1.45 | 2.38 | 8.08 \\
   & \method-7B\textsubscript{ICL}                & 1.70 | 2.72 | 7.97 \\
   & \method-7B\textsubscript{RL}                 & 1.42 | 2.33 | 6.54 \\
\midrule
\multicolumn{3}{c}{\textit{Speaker Similarity}} \\
\midrule 
\multirow{9}{*}{\begin{tabular}[c]{@{}c@{}}\textbf{SEED} \\ \textit{test-zh} | \textit{test-en} | \\ \textit{test-hard} \end{tabular}}   
   & Seed-TTS\textsubscript{ICL}~\citep{seedtts} & 0.796 | 0.762 | 0.776 \\ 
   & Seed-TTS\textsubscript{RL}~\citep{seedtts}  & \textbf{0.801} | \textbf{0.766} | \textbf{0.782} \\ 
   & MaskGCT~\citep{maskgct}                     & 0.774 | 0.714 | 0.748 \\ 
   & E2 TTS~\citep{e2tts}                        & 0.730 | 0.710 | - \\ 
   & F5-TTS~\citep{f5tts}                        & 0.741 | 0.647 | 0.713 \\ 
   & CosyVoice 2~\citep{cosyvoice2}              & 0.748 | 0.652 | 0.724 \\
   & CosyVoice 2-S~\citep{cosyvoice2}            & 0.753 | 0.654 | 0.732 \\
   & \method-7B\textsubscript{ICL}                & 0.752 | 0.632 | 0.747 \\
   & \method-7B\textsubscript{RL}                 & 0.754 | 0.641 | 0.752 \\
\bottomrule
\end{tabular}
\label{tab:zero_shot_speech_generation_table}
\end{table}

\begin{table}[t!]
\centering
\caption{Single-Speaker Speech Generation}
\setlength{\tabcolsep}{2.6pt}
\begin{tabular}{@{}clc@{}}
\toprule
\textbf{Datasets} & \textbf{Model} & \textbf{Performance} \\
\midrule
\multicolumn{3}{c}{\textit{Content Consistency}} \\
\midrule 
\multirow{5}{*}{\begin{tabular}[c]{@{}c@{}}\textbf{SEED} \\ \textit{test-zh} | \textit{test-en} | \\ \textit{test-hard} \end{tabular}} 
   & Human                                & \textbf{1.25} | 2.14 | - \\
   & \method\textsubscript{RL}          & 1.30 | 2.33 | 6.54 \\
   & \method\textsubscript{Speaker A}   & 1.29 | 1.86 | 6.59 \\
   & \method\textsubscript{Speaker B}   & 1.37 | 1.89 | 7.25 \\
   & \method\textsubscript{Speaker C}   & 1.30 | 2.13 | \textbf{6.43} \\
   & \method\textsubscript{Speaker D}   & 1.28 | \textbf{1.83} | 7.16 \\
\midrule
\multicolumn{3}{c}{\textit{Naturalness}} \\
\midrule 
\multirow{5}{*}{\begin{tabular}[c]{@{}c@{}}\textbf{NMOS} \\ \textit{zh} | \textit{en} \end{tabular}} 
   & Human                                & \textbf{4.51} | -     \\
   & \method\textsubscript{Speaker A}   & 4.46 | 4.51 \\
   & \method\textsubscript{Speaker B}   & \textbf{4.51} | \textbf{4.62} \\
   & \method\textsubscript{Speaker C}   & 4.50 | 4.60 \\
   & \method\textsubscript{Speaker D}   & 4.48 | 4.58 \\
\bottomrule
\end{tabular}
\label{tab:zero_shot_speech_generation_table}
\end{table}

\subsubsection{Evaluation of Zero-Shot Speech Generation.}
We compared the \method with state-of-the-art zero-shot TTS systems. As shown in Table 9, \method demonstrates highly competitive performance, highlighting its robust speech understanding and generation capabilities developed through in-context learning (ICL). Additionally, after reinforcement learning (RL) optimization, \method showed significant improvements in generation stability, with marked reductions in attention misalignment, pronunciation errors, and inappropriate pauses on the challenging test-hard dataset.

\subsubsection{Evaluation of Single-Speaker Speech Generation.}
We compared the \method model before and after speaker fine-tuning, as well as with human recordings. As shown in Table 10, the speaker-finetuned \method more precisely captured the nuanced prosodic styles of the target speakers while preserving the foundational stability provided by the base model, achieving performance that approaches human-level quality across both subjective and objective metrics.

\section{Conclusion}
\label{sec:conclusion}
\method is a unified model designed to understand and generate multiple modalities, including text and real-time speech. To enhance video integration, we've introduced a new positional embedding method called TMRoPE, which aligns audio and video timing. Our Thinker-Talker framework supports real-time speech generation while minimizing interference across different modalities. Additionally, we employ techniques such as block-wise audio/vision encoding and a sliding window mechanism for code-to-wav generation. This innovative model excels in complex audio-visual interactions and emotional context in speech dialogues. Comprehensive evaluations show that \method outperforms similarly sized single-modality models, particularly in following voice commands, and achieves state-of-the-art performance in multi-modal tasks.

In the development of the model, we have identified several critical issues that have often been overlooked by researchers in previous academic studies, such as video OCR and audio-video collaborative understanding. Addressing these challenges necessitates collaboration between the academic and industrial sectors, particularly in building comprehensive evaluation benchmarks and research datasets. We believe \method represents a significant advancement toward artificial general intelligence (AGI). Our future goals include developing a more robust and faster model with expanded output capabilities across various modalities like images, videos, and music.

\section{Authors}
\textbf{Core Contributors:} Jin Xu, Zhifang Guo, Jinzheng He, Hangrui Hu, Ting He, Shuai Bai, Keqin Chen, Jialin Wang, Yang Fan, Kai Dang, Bin Zhang, Xiong Wang, Yunfei Chu, Junyang Lin

\textbf{Contributors\footnote{Alphabetical order.}:} An Yang, Anfeng Li, Baosong Yang, Bei Chen, Bin Lin, Binyuan Hui, Bo Zheng, Bowen Yu, Cheng Chen, Chengen Huang, Chenhan Yuan, Chengyuan Li, Daren Chen, Dayiheng Liu, Dake Guo, Fan Zhou, Fei Huang, Guangdong Zhou, Hang Zhang, Haoran Lian, Haoyang Zhang, He Wang, Humen Zhong, Jian Yang, Jiandong Jiang, Jianhong Tu, Jianqiang Wan, Jianyuan Zeng, Jun Tang, Jianwei Zhang, Jianxin Yang, Jianyuan Zeng, Jing Zhou, Jingren Zhou, Kexin Yang, Lei Xie, Linhan Ma, Lingchen Meng, Le Yu, Mei Li, Miao Hong, Mingfeng Xue, Mingkun Yang, Mingze Li, Na Ni, Pei Zhang, Peiyang Zhang, Peng Liu, Peng Wang, Peng Zhang, Pengfei Wang, Rui Hu, Rui Men, Qiuyue Wang, Qing Fu, Shixuan Liu, Sibo Song, Siqi Zhang, Song Chen, Tianyi Tang, Tao He, Ting He, Wenbin Ge, Wei Ding, Xiaodong Deng, Xinyao Niu, Xipin Wei, Xue Bin, Xuejing Liu, Xingzhang Ren, Xuancheng Ren, Yang Liu, Yanpeng Li, Yang Liu, Yang Su, Yichang Zhang, Yuqiong Liu, Yuanjun Lv, Yuanzhi Zhu, Yuxuan Cai, Zeyu Cui, Zheng Li, Zhenru Zhang, Zihan Qiu, Zhaohai Li, Zhibo Yang, Zhipeng Zhou, Zhiyuan Zhu

\bibliography{biblio}
\bibliographystyle{colm2024_conference}

\end{document}